\documentclass[letterpaper, 10 pt, conference]{ieeeconf}  % Comment this line out if you need a4paper

\IEEEoverridecommandlockouts                              % This command is only needed if 

\usepackage{cite}
\usepackage{amsmath,amssymb,amsfonts}
\usepackage[graphicx]{realboxes}
\usepackage{textcomp}
\usepackage[table,xcdraw]{xcolor}
\usepackage{multirow}
\usepackage{array}
\usepackage{tikz}
\usepackage{siunitx}
\usepackage{lscape}
\usepackage{adjustbox}
\usepackage{tabularx}
\usepackage{mleftright}
\usepackage{etoolbox}
\usepackage{xcolor}
\usepackage{amsmath}
\usepackage{amssymb}
\usepackage[normalem]{ulem}
\usepackage{tablefootnote}
\usepackage{verbatim}
\usepackage{graphicx}
\usepackage{booktabs}
\usepackage{mathtools}
\usepackage{makecell}

\usepackage{algorithm}
\usepackage{algpseudocode}

\algtext*{EndFor}   % Removes "end for"
\algtext*{EndIf}    % Removes "end if"
\algtext*{EndWhile} % Removes "end while"
\algtext*{EndFunction} % Removes "end function"

\title{\LARGE \bf
CPR: Mitigating Large Language Model Hallucinations \\
with Curative Prompt Refinement }

\author{Jung-Woo Shim$^{1}$, Yeong-Joon Ju$^{1}$, Ji-Hoon Park$^{1}$, and Seong-Whan Lee$^{1}$
\thanks{*This work was supported by the Institute of Information \& Communications Technology Planning \& Evaluation (IITP) grant, funded by the Korea government (MSIT) (No. RS-2019-II190079, Artificial Intelligence Graduate School Program (Korea University)) and was partly supported by the Institute of Information \& Communications Technology Planning \& Evaluation (IITP) grant, funded by the Korea government (MSIT) (No. RS-2022-II220984, Development of Artificial Intelligence Technology for Personalized Plug-and-Play Explanation and Verification of Explanation).}
\thanks{$^{1}$J.-W. Shim, Y.-J. Ju, J.-H. Park, and S.-W. Lee are with the Department of Artificial Intelligence, Korea University, Anam-dong, Seongbuk-ku, Seoul 02841, Korea.
    {\tt\small j\_w\_shim@korea.ac.kr, yj\_ju@korea.ac.kr, jhoon\_park@korea.ac.kr, and sw.lee@korea.ac.kr}}
}

\begin{document}

\maketitle
\thispagestyle{empty}
\pagestyle{empty}

%%%%%%%%%%%%%%%%%%%%%%%%%%%%%%%%%%%%%%%%%%%%%%%%%%%%%%%%%%%%%%%%%%%%%%%%%%%%%%%%
\begin{abstract}

Recent advancements in large language models (LLMs) highlight their fluency in generating responses to diverse prompts. However, these models sometimes generate plausible yet incorrect ``hallucinated" facts, undermining trust. A frequent but often overlooked cause of such errors is the use of poorly structured or vague prompts by users, leading LLMs to base responses on assumed rather than actual intentions.
To mitigate hallucinations induced by these ill-formed prompts, we introduce Curative Prompt Refinement (CPR), a plug-and-play framework for curative prompt refinement that 1) cleans ill-formed prompts, and 2) generates additional informative task descriptions to align the intention of the user and the prompt using a fine-tuned small language model.
When applied to language models, we discover that CPR significantly increases the quality of generation while also mitigating hallucination. Empirical studies show that prompts with CPR applied achieves over a 90\% win rate over the original prompts without any external knowledge.

\end{abstract}


%%%%%%%%%%%%%%%%%%%%%%%%%%%%%%%%%%%%%%%%%%%%%%%%%%%%%%%%%%%%%%%%%%%%%%%%%%%%%%%%
\section{INTRODUCTION}

While the recent advancements of artificial intelligence span various domains~\cite{arousal2022, pilotEEG2020, 3Dface2020, pillID2012, motion2015}, large language models (LLMs) have displayed unprecedented capabilities in natural language processing and generation. These models, built on sophisticated neural network architectures with notable examples such as ChatGPT~\cite{instructGPT2022} and GPT-4~\cite{GPT42023}, demonstrate remarkable performance in understanding context, answering queries, and creating content that imitates human-like interactions~\cite{humanLike2024}. Despite these advancements, the practical utility of LLMs is often limited due to their generation of hallucinatory content that is plausible yet incorrect, a significant concern highlighted in recent studies~\cite{LLMsurvey2023}. This challenge undermines the reliability of LLM outputs and their applicability across various domains.

\begin{figure}[t!]
\centering
\scriptsize
\centerline{\includegraphics[scale=1]{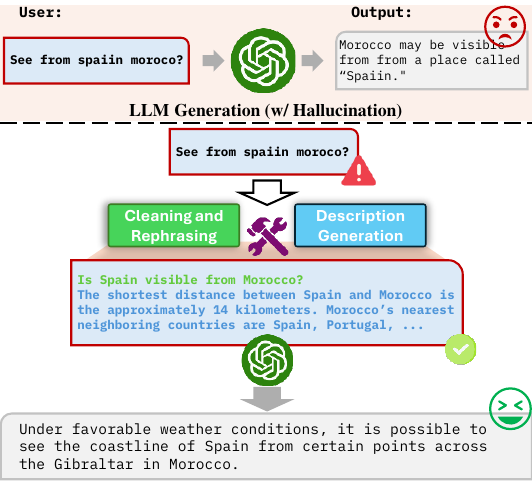}}
\caption{An example of our framework of refining prompts and generating informative task descriptions. With the ill-formed prompt as the input, the GPT-3.5 API generates a hallucinatory output whereas the refined prompt with informative task descriptions generate a high quality response.}
\end{figure}

Most existing studies focus on mitigating hallucinations by modifying the model internally or correcting generated content post-creation~\cite{internal2023}. However, these approaches largely overlook the quality of user inputs, predominantly addressing hallucinations from already well-formed prompts. This oversight reveals a significant shortfall in current methodologies, particularly in how the quality of inputs influences the accuracy and reliability of LLM outputs.

While there are limited studies that do consider the quality of user input, they still encounter significant challenges. These include high computational costs and extensive time demands, often tied to the use of large models or specific constraints related to model enhancements, such as those from reinforcement learning or dependencies on external knowledge bases~\cite{RLHallu2024}. 
These limitations suggest that such studies 
% are not model agnostic and 
heavily rely on additional resources, which complicates their broader applicability and effectiveness.

\begin{figure*}[t] 
  \centerline{\includegraphics[width=1\textwidth]{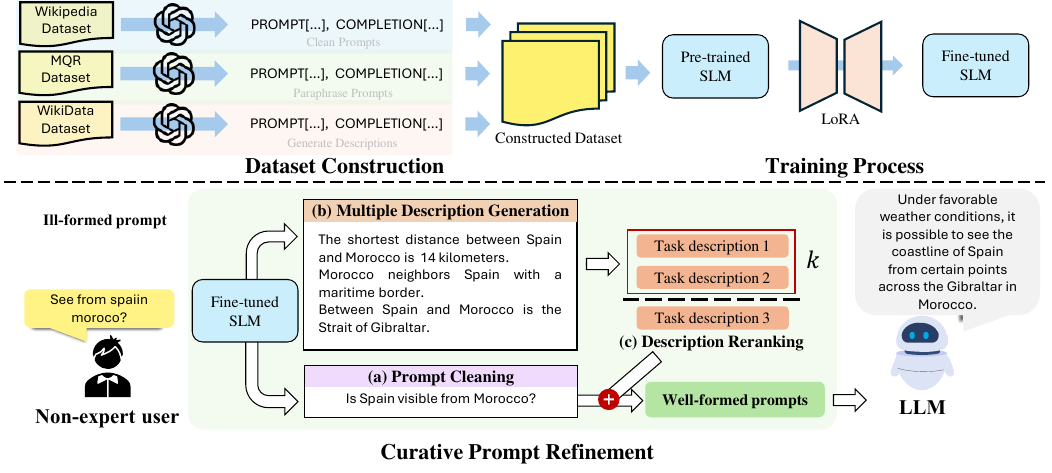}}
  \caption{
  Overview of Curative Prompt Refinement (CPR). 
  First, we fine-tune an SLM on our constructed dataset using LoRA, to mitigate computational burdens in fine-tuning. 
  We then utilize the fine-tuned SLM to (a) refine ill-formed user prompts into prompts without any grammatical errors.
  Following the cleaning process, we (b) generate descriptions of the corresponding prompt, and to maximize the information of the prompts, we (c) use a reranking method to prioritize the most relevant descriptions based perplexity.
  The well-formed prompt allows the inference LLM to generate a concise response.
  } 
  \label{fig} 
\end{figure*}

To address the challenges of ill-formed user inputs and the limitations highlighted in prior research, we introduce Curative Prompt Refinement (CPR), a novel framework designed to refine these inputs before they are processed by LLMs. Our approach specifically targets inputs that often act as out-of-distribution examples, which lead LLMs to generate inaccurate or ``hallucinated" content. 

Initially, we focus on correcting linguistic imperfections to enhance clarity and coherence. We then enrich inputs that lack detailed information by integrating generated, informative task descriptions, ensuring that the prompts are both linguistically accurate and substantively comprehensive. To achieve these goals of prompt refinement and enrichment, we employ a small language model (SLM), noted for its efficiency and cost-effectiveness~\cite{SLM2021}.
Throughout the refinement process, we fine-tune the SLM using three specifically chosen datasets: a paraphrase dataset for word substitution, a general English dataset for punctuation and grammar corrections, and a keyword-description paired dataset for description generation. With capabilities for cleaning prompts and generating descriptions established, we process the prompts and produce the corresponding descriptions following a reranking policy~\cite{smcranking2023}. This policy involves generating multiple descriptions and selecting the top-$k$ descriptions, reranked by perplexity, within a set threshold set upon previous studies~\cite{perplexity2021}. The selected outputs are then combined into a well-structured prompt, thereby reducing the likelihood of the inference LLM generating hallucinations compared to the original, poorly formed prompt.

Our extensive experimental evaluations demonstrate the effectiveness of CPR. The results show that refined prompts, enhanced with detailed task descriptions, significantly improve the quality of outputs from large language models (LLMs) and notably reduce hallucinatory responses. When using CPR, prompts show a 96\% win rate over original, ill-formed prompts with GPT-3.5 as the inference model, and a 99\% win rate over highly ill-formed prompts when a post-processing hallucination mitigation approach is applied.

Moreover, CPR is engineered to be both lightweight and model-agnostic, guaranteeing its applicability across any LLM without tailored adjustments. This semi-universal applicability of our method is a significant advantage, enabling it to integrate as a plug-and-play solution with various LLM architectures. This flexibility, as shown in our experimental results, is particularly beneficial in democratizing access to advanced LLM functionalities, as it evades the need for access to high-end computational resources. By presenting a solution that is both accessible and effective across different computational environments, our approach stands out for its potential to enhance the reliability and quality of LLM outputs for non-expert users without imposing heavy computational or technical prerequisites.

%%%%%%%%%%%%%%%%%%%%%%%%%%%%%%%%%%%%%%%%%%%%%%%%%%%%%%%%%%%%%%%%%%%%%%%%%%%%%%%%
\section{RELATED WORKS}

\subsection{Hallucination Mitigation}

In natural language generation, the phenomenon known as ``hallucination," where models generate plausible but incorrect information, has become a significant area of focus. Efforts to mitigate this issue have predominantly involved refining the internal mechanisms of the models, such as architectural adjustments, augmenting training data, and implementing advanced verification techniques post generation~\cite{snowball2023}. While effective in reducing hallucinations, these methods have limitations in high computational demands and complex implementation processes, making them costly and challenging to scale~\cite{ preventingHallucinations2024}.

\subsection{Prompt Refinement}

Traditional prompt refinement strategies encompass various approaches, including the use of large LLMs, direct human intervention, and reinforcement learning~\cite{PRewrite2024, RewriteLM2024}. These methods typically aim to optimize prompts for a specific target LLM, often encompassing high computational costs. In contrast, our method utilizes a lightweight SLM for prompt refinement, designed to mitigate hallucinations across any LLM. This positions the SLM as a more accessible and resource-efficient option for prompt optimization, achieving high levels of performance through task-specific fine-tuning, despite its smaller scale~\cite{ Distilling2023}. Our approach not only builds on the foundational research in prompt refinement and automated generation~\cite{ smcprompt2023} but also sets itself apart by robustly highlighting the practical advantages of using SLMs.

%%%%%%%%%%%%%%%%%%%%%%%%%%%%%%%%%%%%%%%%%%%%%%%%%%%%%%%%%%%%%%%%%%%%%%%%%%%%%%%%

%%%%%%%%%%%%%%%%%%%%%%%%%%%%%%%%%%%%%%%%%%%%%%%%%%%%%%%%%%%%%%%%%%%%%%%%%%%%%%%%
\section{METHODS}
In this section, we present CPR, a simple yet effective framework for mitigating hallucinations induced by ill-formed prompts.
We describe the details of our framework divided in the following subsections.

\subsection{Dataset Construction}
We first construct a dataset to adapt an SLM to our specific tasks.
Our training dataset for prompt refinement is constructed from three sources: the Wikipedia English dataset (WikiEn)~\cite{Wikipedia2020}, the Multi-domain Question Rewriting dataset (MQR)~\cite{BetterQuestions2020}, and the Wikidata Description dataset (WikiD)~\cite{WikiData2021}. We transform these into instruction fine-tuning form, which are pairs of prompts and ground-truth completions. Each dataset is formatted to train the model for three specific tasks as follows:

\begin{table*}[ht]
\centering
\caption{Comparison of hallucination index (HI), content quality score (CQS), and win rate (WR) between ill-formed of output quality from refined informative prompts upon different inference models. Bold values represent the best values.} 
\renewcommand{\arraystretch}{1.2} 
\setlength{\tabcolsep}{10pt}
\begin{tabular}{lllccc|ccc}
\toprule
\multicolumn{1}{l}{\multirow{3}{*}{\textbf{Method}}} & \multicolumn{1}{c}{\multirow{3}{*}{\textbf{Fine-tuned Models}}} & \multicolumn{6}{c}{\textbf{Inference Models}} \\ 

\multicolumn{1}{c}{} & \multicolumn{1}{c}{} & \multicolumn{3}{c}{\textbf{Llama-2 (7B)~\cite{Llama2023}}} 

& \multicolumn{3}{c}{\textbf{GPT-3.5~\cite{instructGPT2022}}} \\ 
\multicolumn{1}{c}{} & \multicolumn{1}{c}{} & \multicolumn{1}{c}{\textbf{HI} ($\downarrow$)} & \multicolumn{1}{c}{\textbf{CQS} ($\uparrow$)} & \multicolumn{1}{c}{\textbf{WR} ($\uparrow$)} & \multicolumn{1}{c}{\textbf{HI ($\downarrow$)}} & \multicolumn{1}{c}{\textbf{CQS} ($\uparrow$)} & \multicolumn{1}{c}{\textbf{WR ($\uparrow$)}} 

\\ \midrule
\midrule
\multicolumn{1}{l}{Original} & \multicolumn{1}{c}{\textbf{-}} & \multicolumn{1}{c}{0.51} & \multicolumn{1}{c}{0.38} & \multicolumn{1}{c}{\textbf{-}} & \multicolumn{1}{c}{0.16} & \multicolumn{1}{c}{0.51} & \textbf{-} \\ \midrule
\multirow{3}{*}{CPR w/o Descriptions } & Gemma (2B)~\cite{Gemma2024} & \multicolumn{1}{c}{0.35} & \multicolumn{1}{c}{0.51} & \multicolumn{1}{c}{0.51} & \multicolumn{1}{c}{0.14} & \multicolumn{1}{c}{0.58} & 0.61 \\ 
 & Phi-2 (2.7B)~\cite{TinyGPT2023} & \multicolumn{1}{c}{0.34} & \multicolumn{1}{c}{0.54} & \multicolumn{1}{c}{0.53}& \multicolumn{1}{c}{0.12} & \multicolumn{1}{c}{0.63} & 0.64 \\ 
 & Llama-2 (7B)~\cite{Llama2023} & \multicolumn{1}{c}{0.32} & \multicolumn{1}{c}{0.62} & \multicolumn{1}{c}{0.73} & \multicolumn{1}{c}{0.10} & \multicolumn{1}{c}{0.68} & 0.78 \\

 \cmidrule{2-8} 

 & Phi-3 (3.8B) & \multicolumn{1}{c}{{0.15}} & \multicolumn{1}{c}{{0.69}} & \multicolumn{1}{c}{{0.79}} & \multicolumn{1}{c}{{0.08}} & \multicolumn{1}{c}{{0.71}} & {0.81} \\ 
 & Qwen (1.5) (4B) & \multicolumn{1}{c}{0.28} & \multicolumn{1}{c}{0.65} & \multicolumn{1}{c}{0.76} & \multicolumn{1}{c}{0.09} & \multicolumn{1}{c}{0.69} & 0.80 \\ 
 & Zephyr (3B) & \multicolumn{1}{c}{{0.17}} & \multicolumn{1}{c}{{0.68}} & \multicolumn{1}{c}{{0.78}} & \multicolumn{1}{c}{0.09} & \multicolumn{1}{c}{{0.70}} & {0.80} \\ 
 & FLAN-T5 XL (2.85B) & \multicolumn{1}{c}{0.31} & \multicolumn{1}{c}{0.60} & \multicolumn{1}{c}{0.74} & \multicolumn{1}{c}{0.10} & \multicolumn{1}{c}{0.67} & 0.75 \\ 

 \cmidrule{2-8} 

  & Phi-3.5 mini (3B) & \multicolumn{1}{c}{0.13} & \multicolumn{1}{c}{0.77} & \multicolumn{1}{c}{0} & \multicolumn{1}{c}{0} & \multicolumn{1}{c}{0} & 0 \\
   & Llama 3.2 (3.21B) & \multicolumn{1}{c}{0.16} & \multicolumn{1}{c}{0.78} & \multicolumn{1}{c}{0} & \multicolumn{1}{c}{0} & \multicolumn{1}{c}{0} & 0 \\
   
 \midrule
\multirow{3}{*}{CPR (Ours)} & Gemma (2B)~\cite{Gemma2024} & \multicolumn{1}{c}{0.17} & \multicolumn{1}{c}{0.59} & \multicolumn{1}{c}{0.84} & \multicolumn{1}{c}{{0.07}} & \multicolumn{1}{c}{0.68} & 0.92 \\ 
 & Phi-2 (2.7B)~\cite{TinyGPT2023} & \multicolumn{1}{c}{\underline{0.14}} & \multicolumn{1}{c}{{0.63}} & \multicolumn{1}{c}{{0.85}} & \multicolumn{1}{c}{0.09} & \multicolumn{1}{c}{{0.73}} & {0.93} \\ 
 & Llama-2 (7B)~\cite{Llama2023} & \multicolumn{1}{c}{\textbf{0.13}} & \multicolumn{1}{c}{\textbf{0.75}} & \multicolumn{1}{c}{\textbf{0.92}}& \multicolumn{1}{c}{\textbf{0.04}} & \multicolumn{1}{c}{\textbf{0.83}} & \textbf{0.96} \\

 \cmidrule{2-8} 

 & Phi-3 (3.8B) & \multicolumn{1}{c}{\underline{0.14}} & \multicolumn{1}{c}{\underline{0.73}} & \multicolumn{1}{c}{\underline{0.88}} & \multicolumn{1}{c}{\underline{0.05}} & \multicolumn{1}{c}{\underline{0.79}} & \underline{0.94} \\ 
 & Qwen (1.5) (4B) & \multicolumn{1}{c}{0.16} & \multicolumn{1}{c}{0.67} & \multicolumn{1}{c}{0.87} & \multicolumn{1}{c}{0.06} & \multicolumn{1}{c}{0.76} & 0.91 \\ 
 & Zephyr (3B) & \multicolumn{1}{c}{{0.15}} & \multicolumn{1}{c}{{0.71}} & \multicolumn{1}{c}{{0.89}} & \multicolumn{1}{c}{\underline{0.05}} & \multicolumn{1}{c}{{0.77}} & {0.93} \\ 
 & FLAN-T5 XL (2.85B) & \multicolumn{1}{c}{0.19} & \multicolumn{1}{c}{0.65} & \multicolumn{1}{c}{0.85} & \multicolumn{1}{c}{0.07} & \multicolumn{1}{c}{0.74} & 0.88 \\ 

 \cmidrule{2-8} 

  & Phi-3.5 mini (3B) & \multicolumn{1}{c}{0.13} & \multicolumn{1}{c}{\underline{0.77}} & \multicolumn{1}{c}{0} & \multicolumn{1}{c}{0} & \multicolumn{1}{c}{0} & 0 \\
   & Llama 3.2 (3.21B) & \multicolumn{1}{c}{0.16} & \multicolumn{1}{c}{\textbf{0.78}} & \multicolumn{1}{c}{0} & \multicolumn{1}{c}{0} & \multicolumn{1}{c}{0} & 0 \\
   
 \bottomrule
\end{tabular}
\label{table:llm-output-comparison}
\end{table*}

\begin{itemize}
    \item \textbf{WikiEn Dataset}: This dataset comprises a comprehensive collection of English text from Wikipedia that meets high linguistic standards. Using the GPT-3.5 API, we verify and refine 10,000 entries to ensure they are free of punctuation errors, establishing a baseline of well-formed text. Simultaneously, we introduce punctuation errors into a copy of the 10,000 entries using the same API, creating a unique dataset of paired examples. This setup is crucial for fine-tuning the model to improve text correction and punctuation accuracy.
    
    \item \textbf{MQR Dataset}: This dataset, consisting of 2,114 pairs of paraphrased questions, is crucial for enhancing the model's skills in nuanced linguistic transformations like word substitution and paraphrasing. The MQR dataset aids in training the model to transform ill-formed prompts into clearer, more accurate questions, improving its ability to align with user intents and enhance interaction quality and output relevancy.
    
    \item \textbf{WikiD Dataset}: We selected a fraction of the Wikidata dataset based on lookup frequency, comprising 10,000 pairs of keywords and concise descriptions. This selection trains the model to generate accurate descriptions from minimal inputs and to enhance and expand on prompts, producing contextually relevant descriptions. Fine-tuning  an SLM with WikiD equips the model to deliver enriched responses, significantly boosting performance in prompt-based tasks.
\end{itemize}

\subsection{Fine-tuning the Model}
We then fine-tune an SLM with the constructed dataset. For the fine-tuning process, we incorporate instruction fine-tuning techniques~\cite{FLAN2022} coupled with low rank adaptation (LoRA)~\cite{LoRA2022} to enhance parameter efficiency. The approach of utilizing LoRA is specifically chosen to counteract the limitations associated with traditional full fine-tuning, notably its tendency to cause catastrophic forgetting by modifying every parameter of the model~\cite{CatastrophicForgetting2023}. LoRA addresses this by selectively updating only a small subset of model parameters, thereby preserving the pre-trained strengths of the model while still adapting to new tasks.

By refining the model's ability to handle specific instructional tasks without overhauling its entire parameter structure, we achieve a more targeted improvement in performance. This approach not only maintains the general capabilities of the model but also ensures greater efficiency and reduced resource consumption during the fine-tuning phase.

\begin{algorithm}[!t]
\caption{Description Generation and Reranking}
\begin{algorithmic}[1]
\State \textbf{Input:} Input prompt \( q \), maximum number of descriptions \( m \), perplexity threshold \( \tau \), language model \( f \), description \(d\)

\vspace{1mm}
\State \textbf{Define Perplexity Function:}
\vspace{1mm}
\State \( PPL(d_i) = exp[{-\frac{1}{N} \sum_{j=1}^N \log p(w_j|w_{j-1})}] \)
\vspace{1mm}
\Comment{\( N \): tokens in \( d_i \), \( p(w_j|w_{j-1}) \): next-token probability}
\vspace{1mm}
\vspace{1mm}
\State \textbf{Generate and Rerank:}
\State Initialize \( D \) as an empty list
\State \( i = 0 \)
\vspace{1mm}
\While{\( |D| < m \)}
    \State \( d_i \leftarrow f( q ) \) 
    \If{\(PPL(d_i) > \tau \)}
        \State \textbf{break} 
    \Else
        \State Append \( d_i \) to \( D \) 
    \EndIf
    \State \( i \leftarrow i + 1 \)
\EndWhile

\vspace{1mm}
\State \( D \leftarrow \text{sort}(D)\), Set key to \(PPL(x) \) 
\State \( D_{\text{top}} \leftarrow D[:k] \) \Comment{Select top-\( k \) descriptions}

\vspace{1mm}
\State \Return \( D_{\text{top}} \)
\end{algorithmic}
\label{alg1}
\end{algorithm}

\subsection{Cleaning and Paraphrasing Prompts}

Using the fine-tuned SLM, ill-formed prompts are cleaned and rephrased into clear, well-formed versions. The model corrects linguistic inaccuracies such as grammar and syntax errors, a skill developed through extensive training on the Wiki dataset, which includes diverse text examples to enhance error detection and correction. Additionally, it enhances clarity by rephrasing ambiguous phrases, benefiting from its training with the MQR dataset where it learned nuanced linguistic transformations like paraphrasing and word substitution. This dual capability enables the model to deliver prompts that are structurally sound and semantically clear.

\subsection{Informative Description Generation and Reranking}

Merely cleaning the prompt still leaves it lacking in necessary contextual information, which hinders the ability to generate concise responses. This issue occurs because, while SLMs have the necessary information in their corpus, they often struggle to produce it without sufficient contextual guidance.
To mitigate this problem, we enhance prompts by adding supplementary contextual information generated from the model. This process is tailored to align with the refined prompts, improving the LLM's processing effectiveness. Through targeted fine-tuning with our specialized dataset, the SLM learns to generate enriched descriptions that directly complement the cleaned prompts.

To ensure the relevance and quality of the generated descriptions, we continuously monitor the model's perplexity throughout the description generation process. This monitoring persists until the perplexity reaches a predefined threshold of \num{15}. Once this threshold is achieved, we halt the generation process. We then evaluate the descriptions, selecting the top-$k$ based on their average perplexity. These selected descriptions are reranked to prioritize those with the lowest perplexity, ensuring the use of the most coherent and contextually appropriate descriptions. This procedure is outlined in Algorithm~\ref{alg1}.

%%%%%%%%%%%%%%%%%%%%%%%%%%%%%%%%%%%%%%%%%%%%%%%%%%%%%%%%%%%%%%%%%%%%%%%%%%%%%%%%

\section{EXPERIMENTS}

We evaluate CPR in the following aspects: 1) the overall performance of our refined prompts, 2) cleaning and paraphrasing abilities, 3) quality of the generated descriptions, and 4) comparison to an existing hallucination mitigation framework. In the following subsections, we explain the specifics of each experiment, presenting implementation details, evaluation metrics, and comprehensive results.

\subsection{Experimental Settings}
\subsubsection{Evaluation Dataset}To evaluate the effectiveness of CPR to ill-formed prompts we employed a dataset consisting of 8,000 user queries from the Google Well-formed Query dataset~\cite{GWQ2018}, each with a score below 0.5, indicating suboptimal form. To ensure impartiality in processing and evaluation, these queries were anonymized and subjected to a randomization process.
\subsubsection{Hardware Settings}
 Model fine-tuning was conducted using a single NVIDIA RTX A6000 GPU, while inference tasks for the SLMs were performed on a single NVIDIA TITAN V GPU.

\subsection{Evaluating the Effectivness of CPR}
To demonstrate the efficacy of CPR in reducing hallucinations and improving content quality in inference models, we utilized the following metrics evaluated with the GPT-3 API as the judge~\cite{LLMjudge2024}:

\begin{itemize}
\item \textbf{Hallucination Index (HI):} This metric quantifies the accuracy of the content generated by the model, using a scale from 0 to 1. A score of 0 indicates no hallucinations, implying complete factual accuracy, while a score of 1 represents complete hallucination, where the content is entirely fabricated and lacks factual basis.
\item \textbf{Content Quality Score (CQS):} This score evaluates the relevance, coherence, and overall value of the generated content. It is also measured on a scale from 0 to 1, where 0 signifies irrelevant or incoherent content, and 1 indicates relevant and coherent content.
\item \textbf{Win Rate (WR):} This metric measures the superiority of generated content from ill-formed prompts compared to prompts with CPR with no descriptions applied, and ill-formed prompts compared to prompts with CPR applied.  
\end{itemize}

As illustrated in TABLE~\ref{table:llm-output-comparison}, the refined prompts have significantly reduced hallucinations, enhanced content quality, and improved win rates across all tested models, affirming that even small SLMs can produce satisfactory outputs. Our analysis further reveals that the degree of improvement in inference models diminishes as their size increases; smaller models show more substantial improvements. 
Moreover, the comparison of CPR used with 2B (comparably small) and 7B (comparably large) models demonstrates no significant difference in reducing hallucinations when it comes to the size of the SLM, yet it notably enhances quality and win rates. 
This effect is more pronounced when descriptions are included in CPR, where their absence markedly reduces its effectiveness. Interestingly, while the GPT-3.5 model typically performs well, the addition of descriptions through CPR specifically enhances its content quality, indicating the value of context in prompt refining.

\subsection{Evaluation of Prompt Refinement}

To assess the prompt refinement capabilities of the fine-tuned SLMs, we fine-tuned three most widely-used models—Gemma (2B), Phi-2 (2.7B), and Llama-2 (7B)—using our curated dataset. We then evaluated the performance of these models both before and after fine-tuning by comparing the quality of refined prompts against the original ill-formed prompts using the following established metrics:

\begin{itemize}
\item \textbf{BLEU:} This metric measures the grammatical accuracy of translated text, used to assess the quality of refined versus original prompts~\cite{BLEU2002}.
\item \textbf{ROUGE:} This metric evaluates paraphrasing by analyzing n-gram overlap, gauging the fidelity of refined prompts to the originals~\cite{ROUGE2022}.
\item \textbf{METEOR:} This metric assesses semantic similarity using linguistic features, measuring contextual and meaning improvements in refined prompts~\cite{METEOR2005}.
\end{itemize}

By comparing each version of queries before and after refinement, we were able to determine the effectiveness of our dataset in enhancing the fine-tuning process. Our findings are documented in TABLE~\ref{table:refinement-quality}.

\subsection{Evaluation of Generated Descriptions}

To validate the relevance and coherence of descriptions generated by our model, we employed the same fine-tuned SLMs that refined the prompts, since these models were previously adapted for description generation as part of our constructed fine-tuning dataset. For each of the 8,000 queries, the SLMs generated up to five descriptions. Generation was halted when the perplexity of a description reached a predefined threshold of 5, a value determined through empirical research.

\begin{table}[t!]
\caption{Evaluation of prompt cleaning and paraphrasing of SLMs fine-tuned on our dataset. We compare the BLEU, ROUGE, and METEOR scores of each model's generated description and the improvement after fine-tuning. 
}
\centering
\setlength{\tabcolsep}{8pt}
\renewcommand{\arraystretch}{1.2}
\begin{tabular}{clccc}
\toprule
\multirow{1}{*}{\textbf{Model}} 
& \textbf{Fine-tuning} & \textbf{BLEU} & \textbf{ROUGE} & \textbf{METEOR} \\
\midrule
\midrule
\multirow{2}{*}{Gemma~\cite{Gemma2024}}     &Original  & 11.2 & 45.3 & 31.3 \\
& Fine-tuned  & 21.1 & 54.2 & 32.1 \\
\midrule
\multirow{2}{*}{Phi-2~\cite{TinyGPT2023}}    &Original  & 13.1 & 46.6 & 31.7 \\
& Fine-tuned  &\underline{21.7} & \textbf{56.3} & \underline{35.6} \\
\midrule
\multirow{2}{*}{Llama-2~\cite{Llama2023}}    &Original  &12.8 &48.1 &31.5 \\
& Fine-tuned  &\textbf{23.1} &\underline{56.2} &\textbf{36.5} \\
\bottomrule
\end{tabular}
\label{table:refinement-quality}
\end{table}
 
The quality of these descriptions was assessed using the GPT-3 API. We evaluated the descriptions on the following metrics:

\begin{itemize}
\item \textbf{Relevance:} This metric evaluates the extent to which the generated descriptions accurately address the tasks outlined in the prompts. It is assessed on a scale from 0 to 1, where 0 indicates no relevance and 1 represents maximum relevance. Higher scores signify that the content closely aligns with the prompt’s requirements.

\item \textbf{Coherence:} This metric measures the logical consistency and smooth integration of content within the descriptions relative to the refined prompts. It is also rated on a scale from 0 to 1, with 0 signifying illogical content and 1 indicating that the descriptions are perfectly coherent. Higher scores indicates that the model successfully maintains a logical flow and context throughout its responses.
\end{itemize}
% \begin{table}[t!]
% \caption{Comparison of description generation quality of SLMs. We compare relevance and coherence of each model's generated description with the descriptions generated by their fine-tuned models. }
% \centering
% \setlength{\tabcolsep}{15pt}
% \renewcommand{\arraystretch}{1.2} 
% \begin{tabular}{@{}clcc@{}}
% \toprule
% \textbf{Model} & \textbf{Fine-tuning} & \textbf{Relevance} & \textbf{Coherence} \\
% \midrule
% \midrule
% \multirow{2}{*}{Gemma~\cite{Gemma2024}}           & Original & 61.1 & 56.8 \\
%                                 & Fine-tuned & 71.1 & 79.5 \\
% \midrule
% \multirow{2}{*}{Phi-2~\cite{TinyGPT2023}}          & Original & 61.4 & 58.8 \\
%                                 & Fine-tuned & \underline{73.4} & \underline{81.4} \\
% \midrule
% \multirow{2}{*}{Llama-2~\cite{Llama2023}}          & Original & 63.4 & 62.5 \\
%                                 & Fine-tuned & \textbf{80.5} & \textbf{82.1} \\
% \bottomrule
% \end{tabular}
% \label{table:generation-quality}
% \end{table}

\begin{table}[t!]
\caption{Comparison of description generation quality of SLMs. We compare relevance and coherence of each model's generated description with the descriptions generated by their fine-tuned models. }
\centering
\setlength{\tabcolsep}{10pt}
\renewcommand{\arraystretch}{1} 
\begin{tabular}{@{}clcc@{}}
\toprule
\textbf{Model} & \textbf{Fine-tuning} & \textbf{Relevance} & \textbf{Coherence} \\
\midrule
\midrule
\multirow{2}{*}{Gemma~\cite{Gemma2024}}           & Original & 61.1 & 56.8 \\
                                & Fine-tuned & 71.1 & 79.5 \\
\midrule
\multirow{2}{*}{Phi-2~\cite{TinyGPT2023}}          & Original & 61.4 & 58.8 \\
                                & Fine-tuned & {73.4} & {81.4} \\
\midrule
\multirow{2}{*}{Llama-2~\cite{Llama2023}}          & Original & 63.4 & 62.5 \\
                                & Fine-tuned & \textbf{80.5} & \textbf{82.1} \\
\midrule
\multirow{2}{*}{Phi-3 (3.8B)}           & Original & 64.1 & 63.0 \\
                                & Fine-tuned & \underline{78.9} & \underline{81.7} \\
\midrule
\multirow{2}{*}{Qwen (1.5) (4B)}           & Original & 62.5 & 61.0 \\
                                & Fine-tuned & 75.8 & 80.3 \\
\midrule
\multirow{2}{*}{Zephyr (3B)}           & Original & 63.0 & 61.8 \\
                                & Fine-tuned & {77.5} & {81.2} \\
\midrule
\multirow{2}{*}{FLAN-T5 XL (2.85B)}           & Original & 62.0 & 60.5 \\
                                & Fine-tuned & 74.2 & 79.0 \\
                                \midrule
\multirow{2}{*}{Phi-3.5 mini (3B)}           & Original & \textit{67.3} & \textit{66.1} \\
                                & Fine-tuned & \textit{0} & \textit{0} \\
\midrule
\multirow{2}{*}{Llama-3.2 (3.21B)}           & Original & \textit{69.8} & \textit{66.3} \\
                                & Fine-tuned & \textit{0} & \textit{0} \\
\bottomrule
\end{tabular}
\label{table:generation-quality}
\end{table}

Each description was independently reviewed and scored by the GPT-3 API, with results averaged. Our experiment demonstrates that while SLMs initially show limited performance in both prompt refinement and description generation, they can be effectively fine-tuned to achieve satisfactory outcomes. The results are documented in TABLE~\ref{table:generation-quality}.

\subsection{Comparison of CPR with SelfCheckGPT}
To demonstrate the effectiveness of CPR in reducing hallucinations, we conducted a comparative experiment against SelfCheckGPT, one of the current leading technique for hallucination mitigation, with Llama-2 (7B) as its inference model. We assessed both methods using the same metrics—HI, CQS, and WR—with two different models, Gemma (2B) and Llama-2 (7B), serving as the inference models. We specifically chose Gemma (2B), the smallest SLM as CPR's SLM.

As shown in TABLE~\ref{table:baseline}, CPR outperforms SelfCheckGPT when handling highly ill-formed prompts (prompts with scores under 0.2; High). However, for prompts with minimal errors (prompts with scores above 0.2; Low), SelfCheckGPT was more effective. Notably, combining CPR with SelfCheckGPT yielded the highest scores across all metrics, showcasing the compatibility and enhanced performance of integrating these methods. 

\begin{table}[t!]
\caption{Comparison of CPR with SelfCheckGPT~\cite{Selfcheckgpt2023}. We compare the mitigation of hallucination for ill-formed inputs using different SLMs as the inference model. IF represents the ill-formed degree of the prompts.}
\begin{center}
\centering
\renewcommand{\arraystretch}{1.2} 
\begin{tabular}{llccc}
\toprule

\textbf{IF Degree} & \textbf{Method} &  \textbf{HI ($\downarrow$)} & \textbf{CQS ($\uparrow$)} & \textbf{WR ($\uparrow$)} \\
\midrule
\midrule
\multirow{4}{*}{Low}
&SelfCheckGPT~\cite{Selfcheckgpt2023}& \textbf{0.19} & \underline{0.58} & \textbf{0.71} \\
&CPR w/ Gemma (2B)& 0.23 & 0.42 & 0.61 \\
&CPR w/ Llama-2 (7B)& \underline{0.21} & \textbf{0.62} & \underline{0.68} \\
&\cellcolor[HTML]{EFEFEF}CPR+SelfCheckGPT &\cellcolor[HTML]{EFEFEF}0.03 &\cellcolor[HTML]{EFEFEF} 0.91&\cellcolor[HTML]{EFEFEF} 0.98\\
\midrule
\multirow{4}{*}{High}

&SelfCheckGPT~\cite{Selfcheckgpt2023}& 0.37 & 0.42 & 0.51 \\
&CPR w/ Gemma (2B)& \underline{0.29} & \underline{0.48} & \underline{0.61} \\
&CPR w/ Llama-2 (7B)& \textbf{0.23} & \textbf{0.57} & \textbf{0.69} \\
&\cellcolor[HTML]{EFEFEF}CPR+SelfCheckGPT &\cellcolor[HTML]{EFEFEF}0.05 &\cellcolor[HTML]{EFEFEF} 0.88&\cellcolor[HTML]{EFEFEF}0.99 \\
\bottomrule

\end{tabular}
\end{center}
\label{table:baseline}
\end{table}

%%%%%%%%%%%%%%%%%%%%%%%%%%%%%%%%%%%%%%%%%%%%%%%%%%%%%%%%%%%%%%%%%%%%%%%%%%%%%%%%
\section{CONCLUSION}

We introduce CPR, an automatic plug-and-play framework for refining prompts utilizing an SLM, to ultimately mitigate hallucinations.
Our approach involves cleaning, paraphrasing, and generating informative descriptions adequate to the prompts. This substantially improves the clarity and task specificity of prompts, leading to a significant uplift in the quality of LLM outputs as shown in our experimental results.
Despite these advancements, there are still some issues that remain. 
Firstly, the dataset crafted for fine-tuning was manually done, indicating a better crafted dataset potentially could amplify the effectiveness of our approach. 
Additionally, resource constraints restricted our evaluation scope, limiting a thorough exploration of scalability and performance across models with larger sizes. 
Future works will deal with these issues by crafting a higher-quality dataset for fine-tuning and experimenting on a wider range of inference models.
Nonetheless, our findings contribute significantly to the field, underscoring the substantial benefits of incorporating SLMs into the preprocessing stages for LLMs. 
The marked improvements in the accuracy and quality of LLM-generated content highlight the potential of CPR.

%%%%%%%%%%%%%%%%%%%%%%%%%%%%%%%%%%%%%%%%%%%%%%%%%%%%%%%%%%%%%%%%%%%%%%%%%%%%%%%%

\bibliographystyle{IEEEtran}
\bibliography{REFERENCE}
% \begingroup
% \small
% \input{root.bbl}  % must match main file base name
% \endgroup

\end{document}